\documentclass{article}

\usepackage{microtype}
\usepackage{graphicx}
\usepackage{subfigure}
\usepackage{booktabs} 

\usepackage{hyperref}
\usepackage{hyperref,url,booktabs,amsfonts,nicefrac,microtype, xcolor, graphicx,amsmath,floatrow, subfig, wrapfig, enumerate, makecell, multirow, stackengine,wasysym,scalerel, caption, bbm}

\usepackage{subcaption}
\usepackage{caption}

\newlength\savewidth\newcommand\shline{\noalign{\global\savewidth\arrayrulewidth
  \global\arrayrulewidth 1pt}\hline\noalign{\global\arrayrulewidth\savewidth}}

\newcolumntype{x}[1]{>{\centering\arraybackslash}p{#1pt}}
\newcolumntype{y}[1]{>{\raggedright\arraybackslash}p{#1pt}}
\newcolumntype{z}[1]{>{\raggedleft\arraybackslash}p{#1pt}}
\usepackage{float}

\newcommand{\app}{\raise.17ex\hbox{$\scriptstyle\sim$}}

\newfloatcommand{capbtabbox}{table}[][\FBwidth]

\usepackage[accepted]{icml2024}

\usepackage{amsmath}
\usepackage{amssymb}
\usepackage{mathtools}
\usepackage{amsthm}

\usepackage[capitalize,noabbrev]{cleveref}

\theoremstyle{plain}

\theoremstyle{definition}

\theoremstyle{remark}

\expandafter\def\expandafter\normalsize\expandafter{%
    \normalsize%
    \setlength\abovedisplayskip{0pt}%
    \setlength\belowdisplayskip{8pt}%
    \setlength\abovedisplayshortskip{-8pt}%
    \setlength\belowdisplayshortskip{2pt}%
}

\usepackage[textsize=tiny]{todonotes}

\icmltitlerunning{NEBULA}

\newcommand{\TODO}[1]{{}} 

\begin{document}

\twocolumn[

\icmltitle{NEBULA: \underline{N}eural \underline{E}mpirical \underline{B}ayes \underline{U}nder \underline{LA}tent Representations for Efficient and Controllable Design of Molecular Libraries}




\begin{icmlauthorlist}
\icmlauthor{Ewa M. Nowara}{pd-sf}
\icmlauthor{Pedro O. Pinheiro}{pd-nyc}
\icmlauthor{Sai Pooja Mahajan}{pd-sf}
\icmlauthor{Omar Mahmood}{pd-nyc}
\icmlauthor{Andrew Martin Watkins}{pd-sf}
\icmlauthor{Saeed Saremi}{pd-sf}
\icmlauthor{Michael Maser}{pd-sf}

\end{icmlauthorlist}

\icmlaffiliation{pd-nyc}{Prescient Design, Genentech, New York City, USA}
\icmlaffiliation{pd-sf}{Prescient Design, Genentech, South San Francisco, USA}

\icmlcorrespondingauthor{Ewa M. Nowara}{nowara.ewa@gene.com}

\icmlkeywords{Machine Learning, Drug Discovery, Generative Models, 3D Generation, Voxel Structures, Molecules}

\vskip 0.3in
]



\printAffiliationsAndNotice{}  

\begin{abstract}
We present NEBULA, the first latent 3D generative model for scalable generation of large molecular libraries around a seed compound of interest. Such libraries are crucial for scientific discovery, but it remains challenging to generate large numbers of high quality samples efficiently. 3D-voxel-based methods have recently shown great promise for generating high quality samples \textit{de novo} from random noise~\cite{pinheiro20233d}. However, sampling in 3D-voxel space is computationally expensive and use in library generation is prohibitively slow. Here, we instead perform \textit{neural empirical} Bayes sampling~\cite{saremi2019neural} in the learned latent space of a vector-quantized variational autoencoder. NEBULA generates large molecular libraries nearly \textbf{an order of magnitude faster} than existing methods without sacrificing sample quality. Moreover, NEBULA generalizes better to unseen drug-like molecules, as demonstrated on two public datasets and multiple recently released drugs. We expect the approach herein to be highly enabling for machine learning-based drug discovery. The code is available at \href{https://github.com/prescient-design/nebula}{https://github.com/prescient-design/nebula}.
\end{abstract}

\section{Introduction}
\label{introduction}

Computational generation of new molecules with desired properties is a key element of chemical research, especially in drug discovery (DD). 
While search-based methods have achieved some success~\cite{ghorbani2023autoregressive, kowalski2023automated, janda1994tagged}, they can only explore small portions of chemical space, which is often insufficient for modern DD. Machine learning (ML) generative models can potentially accelerate search by rapidly generating large molecular libraries closely resembling a seed molecule of interest~\cite{Bilodeau}. However, doing so while maintaining high sample quality remains an open challenge.

Chemical space exploration for the \textit{optimization} of molecular properties remains at the forefront of DD. Traditional experimental approaches can be slow and unsatisfactory in results, in part due to the economic infeasibility of examining large enough chemical spaces in the wet lab~\cite{wang2024present}. Methods for the prediction of molecules' biological properties have enabled more experimentation to occur \textit{in silico} and have increased the rate of discovery and triage~\cite{konze2019reaction, khalak2022chemical, thompson2022optimizing, maser2023supsiam}. However, these are still largely dependent on \textit{ad hoc}, combinatorial, and/or commercially available definitions of search space, which are unlikely to yield precise compounds with globally optimal properties. Therefore, combined with property predictors, expanding virtual libraries (VLs) beyond such definitions with generative models could be highly enabling for MLDD~\cite{wang2024present}.

Molecules exist in 3D space, and 3D representations, such as volumes (voxels), capture rich information about their structure, shape, and properties. 3D generative models (3DGMs) trained with such inputs~\cite{hoogeboom2022equivariant, pinheiro20233d} have the potential to learn more complete representations of molecules compared to models trained with 1D sequences~\cite{Segler, Blaschke, guimaraes2018objectivereinforced} or 2D graphs~\cite{pmlr-v80-jin18a, li2018learning, you2018graphrnn, Mahmood}.
However, generative sampling in discrete 3D-input space is very computationally intensive~\cite{pinheiro20233d}, making it impractical for creating VLs around lead chemical matter in which to perform optimization. 

\TODO{explain that we need large libraries bc most of small molecules that will be proposed may not be synthesizable?}

\TODO{move this to Related Work?}

We herein present an approach to overcome such challenges by training denoising autoencoders (DAEs) on the continuous, low-dimensional latent representations of 3DGMs, which we call NEBULA. We demonstrate that \textit{neural empirical} Bayes (NEB) sampling~\cite{saremi2019neural} around noisy latent vectors substantially reduces the cost \textit{and} increases control of sample generation in the neighborhood surrounding input seed molecules, which we find highly desirable. Further, we show that training under our approach improves generalizability to chemical spaces well outside of training distributions, including for building-block and fragment generation. Finally, we provide real-world examples of candidate VLs generated around recently disclosed small-molecule drugs.

The main contributions of this work are the following:
\begin{enumerate}
\itemsep0.01em 
\item Our latent approach, NEBULA, generates new molecules up to an order of magnitude faster than the state-of-the-art (SOTA) 3DGMs~\cite{pinheiro20233d, xu2023geometric} while maintaining high sample quality;
\item NEBULA scales to generation of very large molecular libraries with stable, unique, and valid (SUV) molecules similar to a seed compound;
\item Finally, NEBULA generalizes better than SOTA to generation around molecules from new chemical spaces, including multiple examples of real drugs that were just released in March 2024. 
\end{enumerate}

\begin{figure*}[t]
\vskip 0.2in
\centering
\centerline{\includegraphics[width=\columnwidth]{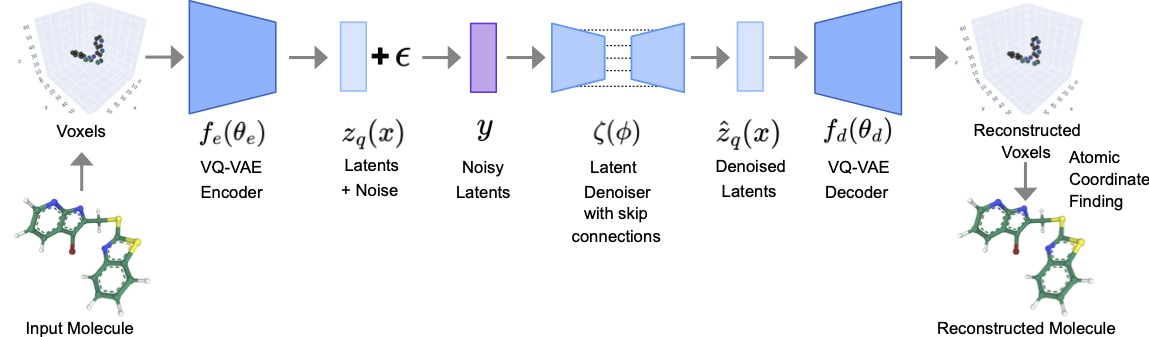}}
\caption{Overview of the proposed latent generative model, NEBULA. A 3D molecular graph is represented as a voxel grid and is passed through a VQ-VAE encoder to obtain latent embeddings. Noise is added and denoised by a latent U-Net, which is used to generative sampling. Denoised latents are passed through a VQ-VAE decoder to reconstruct the voxel grid, and the molecular graph is obtained via peak finding and sanitization.}
\label{fig:overview}
\vskip -0.2in
\end{figure*}

\section{Related Work}
\label{related_work}

\subsection{Molecular Library Generation}
The majority of modern methods for large molecular-library generation are based on non-ML approaches involving cheminformatics or combinatorial enumeration~\cite{janda1994tagged, konze2019reaction, ghorbani2023autoregressive, kowalski2023automated}. These methods are very often restricted to molecules within a commercially available chemical space. This is a significant limitation as most advanced DD programs require exploring new molecules outside of the commercially available space, which ML generative methods excel at as they able to produce completely novel samples.

\subsection{Generation of Molecules in 3D Space}
Generation of molecules represented in 3D space has been successful in creating high quality molecules. \citet{wang2022pocket} used a generative adversarial network~\cite{goodfellow2014generative} on voxelized electron densities for pocket-based generation of molecules resembling known ligands. \citet{skalic2019shape} and
\citet{ragoza2020learning} generated 3D voxelized molecules with CNNs and variational autoencoders (VAEs)~\cite{kingma2013auto}. VoxMol~\cite{pinheiro20233d} similarly generates molecules as 3D voxels with a \textit{walk-jump sampling} (WJS) generative approach~\cite{saremi2019neural}, which currently holds SOTA performance in terms of sample quality and efficiency of generation. GSchNet~\cite{gebauer2019symmetry} used autoregressive models to iteratively sample atoms and bonds in 3D space. \citet{hoogeboom2022equivariant} proposed E(3) Equivariant Diffusion Model (EDM) that learns to generate molecules by iteratively applying a denoising network to a noise initialization. Finally, \citet{vignac2023midi} improved EDM by jointly generating the 3D conformation and the 2D connectivity graph of molecules.

\subsection{Latent-based Generation} 
Latent diffusion models (LDMs)~\cite{rombach2022high} have enabled efficient generation of 2D images by compressing input data to a learned latent space where denoising and sampling is less computationally intensive. \citet{xu2023geometric} applied EDM in the latent space instead of the input atomic coordinates to generate molecular \textit{geometries}, dubbed GeoLDM. Finally, \citet{mahajanexploiting} demonstrated the effectiveness of NEB sampling in the latent space of pre-trained language models for generation of novel 1D protein sequences. 

It is worth noting that EDM and other geometric methods require node matrices of a predetermined size for generation; i.e., the number of atoms $N$ in a generated sample must be given. This is a significant limitation for our setting of seeded generation, where $N$ may take on a large array of values around an input molecule with $N_0$, especially when, e.g., adding a new fragment or functional group. We avoid this limitation by leveraging 3D voxel representations, while still retaining the advantages of latent-based generation and NEB. Our work is the first latent-based 3DGM of voxelized molecular denisties, and the first 3DGM suitable for library generation relative to lead chemical matter.

\section{Proposed Approach}
We present, NEBULA, \underline{N}eural \underline{E}mpirical \underline{B}ayes \underline{U}nder \underline{LA}tent Representations. NEBULA is an efficient latent score-based generative model which compresses voxelized molecules to a lower dimensional latent space and generates new molecules by sampling around noised latent embeddings using NEB~\cite{saremi2019neural}. See Figure~\ref{fig:overview} for an overview of our proposed framework. 

\subsection{Compressing Voxels to a Latent Representation}

We obtain 3D voxel representations of molecules by treating each atom as a continuous Gaussian density in 3D space centered around its atomic coordinates on a voxel grid by following~\citet{pinheiro20233d}. Each molecule is represented as a 4D tensor of $[c \times l \times l \times l]$, where $c$ is the number of atom types and $l$ is the length of the voxel grid edge. The values of each voxel range between 0 and 1.

Following common practice~\cite{rombach2022high, dai2019diagnosing, esser2021taming, ramesh2021zero, razavi2019generating, van2017neural}, we separate the training of a compression model to learn a latent representation and the training of a latent denoising model used for generation in two independent stages. As in Stable (latent) Diffusion (LDMs), we train a vector-quantized VAE (VQ-VAE)~\cite{van2017neural} to compress voxels to a latent space and then decode back to the original input. VQ-VAEs have been shown to aid in producing stable and valid samples for images~\cite{rombach2022high} and molecules~\cite{wang2022pocket}. The input voxelized molecules $x$ are encoded with a network $f_{e}(x)$ parameterized by $\theta_e$ to continuous latent embeddings $z_{e}$ (Eq.~\ref{eq:1}). Each latent dimension is quantized to discrete vector $z_q$ by matching it with one of $k$ vectors in a learned ``codebook'' of embeddings $e$ via nearest neighbor look-up (Eq.~\ref{eq:2}). The quantized latent embeddings $z_{q}$ are then passed through the decoder $f_d$ ($\theta_d$) to reconstruct the input voxels $\hat{x}$ (Eq.~\ref{eq:3}). 

\begin{equation} 
f_{e}(x): x \rightarrow z_{e} 
\label{eq:1}
\end{equation}

\begin{equation}
z_{q} \leftarrow e_{k}, \textrm{where } k=\textrm{argmin}_{j} ||z_{e} - e_{j}||_{2}
\label{eq:2}
\end{equation}

\TODO{fix j under argmin}

\begin{equation}
f_{d}(z_q): z_{q} \rightarrow \hat{x} 
\label{eq:3}
\end{equation}

The VQ-VAE is trained with a loss (Eq.~\ref{eq:4}) comprised of 1) a mean-squared error (MSE) reconstruction term between the input $x$ and reconstructed voxels $\hat{x}$, 2) loss to learn the codebook of the embeddings $e$ by moving the learned quantized embedding vectors $e{_i}$ towards the continuous latent embeddings $z_{e}$, 3) and a ``commitment cost" term which ensures the outputs of the encoder  do not grow arbitrarily, where $\beta$ is a hyperparameter and ``sg" denotes a \texttt{stopgradient} operation.

\begin{equation}
    \mathcal{L}_V = ||\hat{x} - x ||_{2}^{2} + ||\textrm{sg}[z_{e}] - e||_{2}^{2} + \beta ||z_{e} - \textrm{sg}[e]||_{2}^{2}
    \label{eq:4}
\end{equation}

\TODO{move to appendix if needed}

\subsection{Denoising Latent Embeddings}
For generation, we train a DAE on the learned VQ-VAE latents by adding isotropic Gaussian noise to the quantized embeddings $z_{q} \rightarrow z_q^{\prime}$ with identity covariance matrix scaled by a fixed large noise level $\sigma$ (Eq.~\ref{eq:5}). We normalize the latent embeddings before adding noise by subtracting the mean $\mu_c$ and dividing by the standard deviation $\sigma_c$ computed across the $[c * l * l * l]$ channels over the training set (the empirical latent posterior). We train the latent model $\zeta_{\phi}$ to denoise $z_q^{\prime}$ with a MSE loss computed between the input and predicted embeddings $z_{q}$ and $\hat{z}_q$, respectively (Eq.~\ref{eq:6}). The denoised latents are then unnormalized with $\mu_c, \sigma_c$ before passing them to the decoder $f_d$ for voxel reconstruction.


\begin{equation}
z_q^{\prime} = z_{q} + \epsilon, \epsilon \sim \mathcal{N}(0, \sigma^{2} I_{d}) 
\label{eq:5}
\end{equation}

\begin{equation}
\mathcal{L_D} = ||z_{q} - \hat{z}_{q}||_{2}^{2} \: ; \: \hat{z}_{q} = \zeta_\phi(z_q^{\prime})
\label{eq:6}
\end{equation}

\subsection{Generation with Walk Jump Sampling}

We generate new molecules with a \textit{neural empirical} Bayes sampling scheme~\cite{saremi2019neural} performed in the latent space in a two step process, referred to as \textit{walk-jump sampling} (WJS). The WJS generative approach is a score-based method similar to diffusion models but it is more efficient as it only requires a single denoising step and noise scale compared to iterative denoising and complex noise schedules required in DMs. WJS is well suited for structured and textureless data, such as molecules and amino acid sequences~\cite{frey2023learning}, but it has also been successfully used for generation of images~\cite{saremi2022multimeasurement, saremi2023universal}. First, we sample new noisy latent embeddings from the smooth (noisy) latent distribution using Langevin Markov chain Monte Carlo (MCMC) sampling~\cite{cheng2018underdamped} with multiple $k$ \textit{walk steps} along the initialized manifold, where $B_t$ is standard Brownian motion, $\gamma$ and $u$ are hyperparameters (referred to as friction and inverse mass, respectively), and $g_\zeta$ is the learned latent denoising score function.

\begin{equation}
\begin{split}
\begin{aligned}
d\upsilon_t = - \gamma \upsilon_t dt - ug_\zeta (z_{qt}^{\prime}) dt + (\sqrt{2\gamma u}) dB_t 
\\
d z_{qt}^{\prime} = \upsilon_t dt
\label{eq:7}
\end{aligned}
\end{split}
\end{equation}

Second, we denoise the latent embeddings in a \textit{jump step} with a forward pass of the denoising model for an arbitrary step $k$.

\begin{equation}
z_{qk} = \zeta_\phi(z_{qk}^{\prime})
\label{eq:8}
\end{equation}

We obtain the generated molecules by reconstructing the 3D voxels from the newly generated latent embeddings by passing them through the decoder of the frozen compression VQ-VAE model. Finally, we obtain the molecular graphs from the voxels by using the approach in~\citet{pinheiro20233d}, which is an optimization-based peak finding used to locate the atomic coordinates as the center of each voxel.

\section{Data and Methods}\label{sec:data_and_methods}
\textbf{Datasets.} We train NEBULA on GEOM-drugs (GEOM)~\cite{geomdrugs_axelrod2022geom}, which is a standard dataset used for molecule generation. We follow~\citet{vignac2023midi} to split it into the train, validation, and test sets with 1.1M/146K/146K molecules each. We use soft random subsampling~\cite{cui2023soft} and train on 10\% of the training set in each epoch. We test the cross-dataset generalizability of NEBULA and VoxMol trained on GEOM to a different large public dataset called PubChem Quantum (PCQM)~\cite{pubchem1,pubchem2,pubchem3} and only use its test set containing 10,000 molecules. We additionally test the generalizability of NEBULA to several real drug molecules disclosed earlier this year~\cite{ACS}. We use voxel grids of dimension 64, 8 atom channels ([C, H, O, N, F, S, Cl, Br]), and atomic radii of 0.25 \r{A} resolution for all experiments.  

\textbf{Training Details.} We use a VQ-VAE 3D autoencoder with no skip connections as the compression model architecture and a 3D U-Net with skip connections for the latent denoiser. Both models use a 3D convolutional architecture similar to~\cite{pinheiro20233d}, with 4 levels of resolution and self-attention on the lowest two resolutions. We found that a latent dimension of 1024 worked best with higher noise levels. We train the compression and latent denoising models for about 150 epochs. We augment all models during training by randomly rotating and translating every training sample. We use a noise level of $\sigma=1.8$ for all generations. See Appendix for details of the architecture, architecture ablations, training and sampling. 

\textbf{Baselines.} The majority of existing ML generative models for molecules were only evaluated on \textit{de novo} generation from random noise~\cite{gebauer2019symmetry, hoogeboom2022equivariant, pinheiro20233d}. Therefore, to compare our method to existing approaches, we implemented and evaluated VoxMol~\cite{pinheiro20233d} for generation around a seed molecule, as it is the SOTA molecule generation method. Similar to NEBULA, VoxMol is a voxel-based 3D CNN approach, however, it generates molecules in the large input voxel space.


\textbf{Evaluation Metrics.} 
We evaluate the quality of the generated molecules with multiple metrics used by~\citet{vignac2023midi} and~\citet{pinheiro20233d}, such as molecular stability and validity (details in the Appendix). We also quantify how similar the generated molecules are to the seed molecule with the Tanimoto similarity computed between the seed and the generated molecular graphs~\cite{Bender2004MolecularSA}. Following~\cite{vignac2023midi} we report the results by only keeping the largest generated molecule in cases where multiple molecule fragments are generated for each generation. 

\section{Results and Discussion}
\subsection{Molecular Library Generation}\label{sec:molecular_library_generation}

\begin{figure}
    \vskip 0.2in
    \centering
    \begin{subfigure}
        \centering
        \includegraphics[scale=0.26]{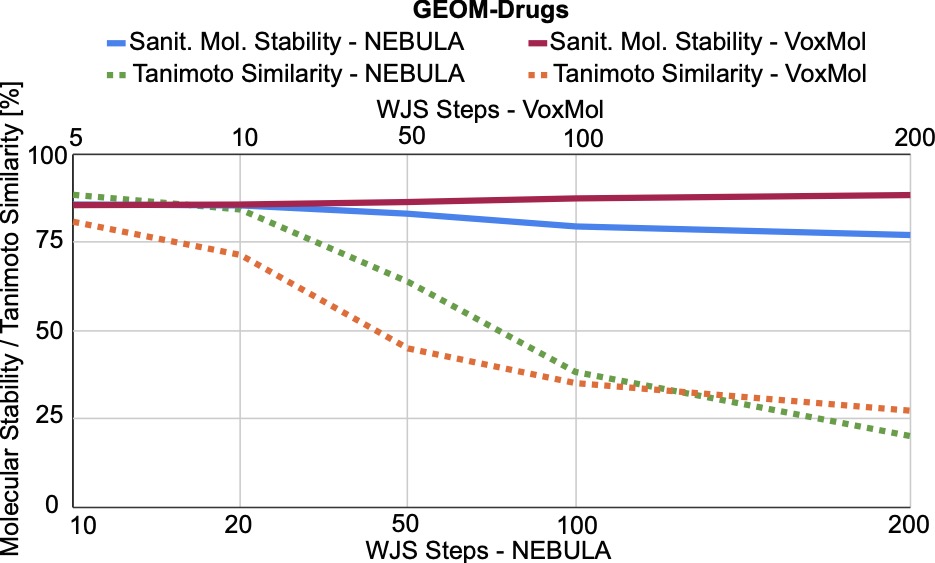}
    \end{subfigure}
    
    \begin{subfigure}
        \centering
        \includegraphics[scale=0.26]{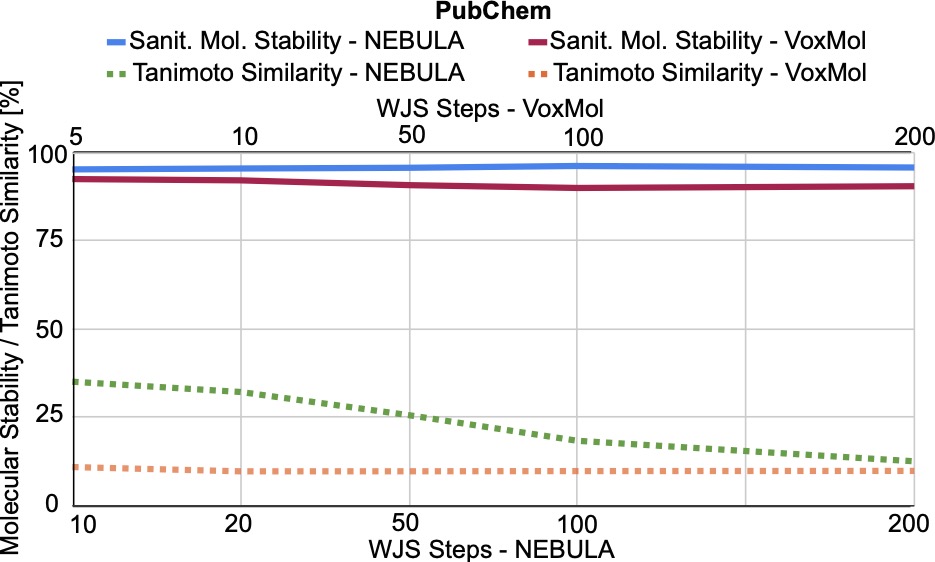}
    \end{subfigure}

    \begin{subfigure}
        \centering
        \includegraphics[scale=0.26]{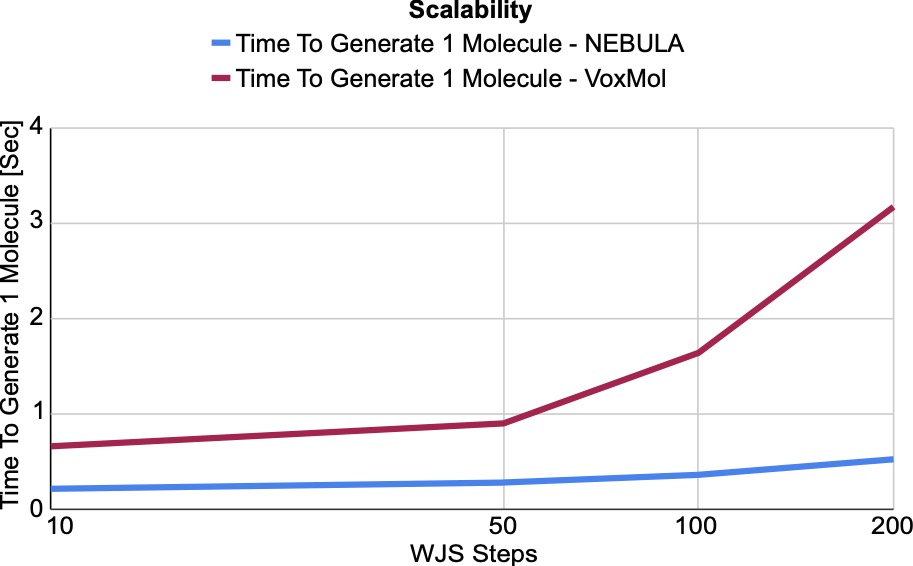}
    \end{subfigure}
    
    \caption{Molecular stability and Tanimoto similarity over WJS steps for molecules generated with NEBULA and VoxMol on \textbf{GEOM} (top) and \textbf{PCQM} (middle). (bottom) Scalability of each method plotted as the amount of time needed to generate one molecule at different WJS steps.}
    \vskip -0.2in
    \label{fig:stab_tan_time}
\end{figure}

\begin{figure*}
\vskip 0.2in
\centering
    \includegraphics[width=0.98\textwidth]{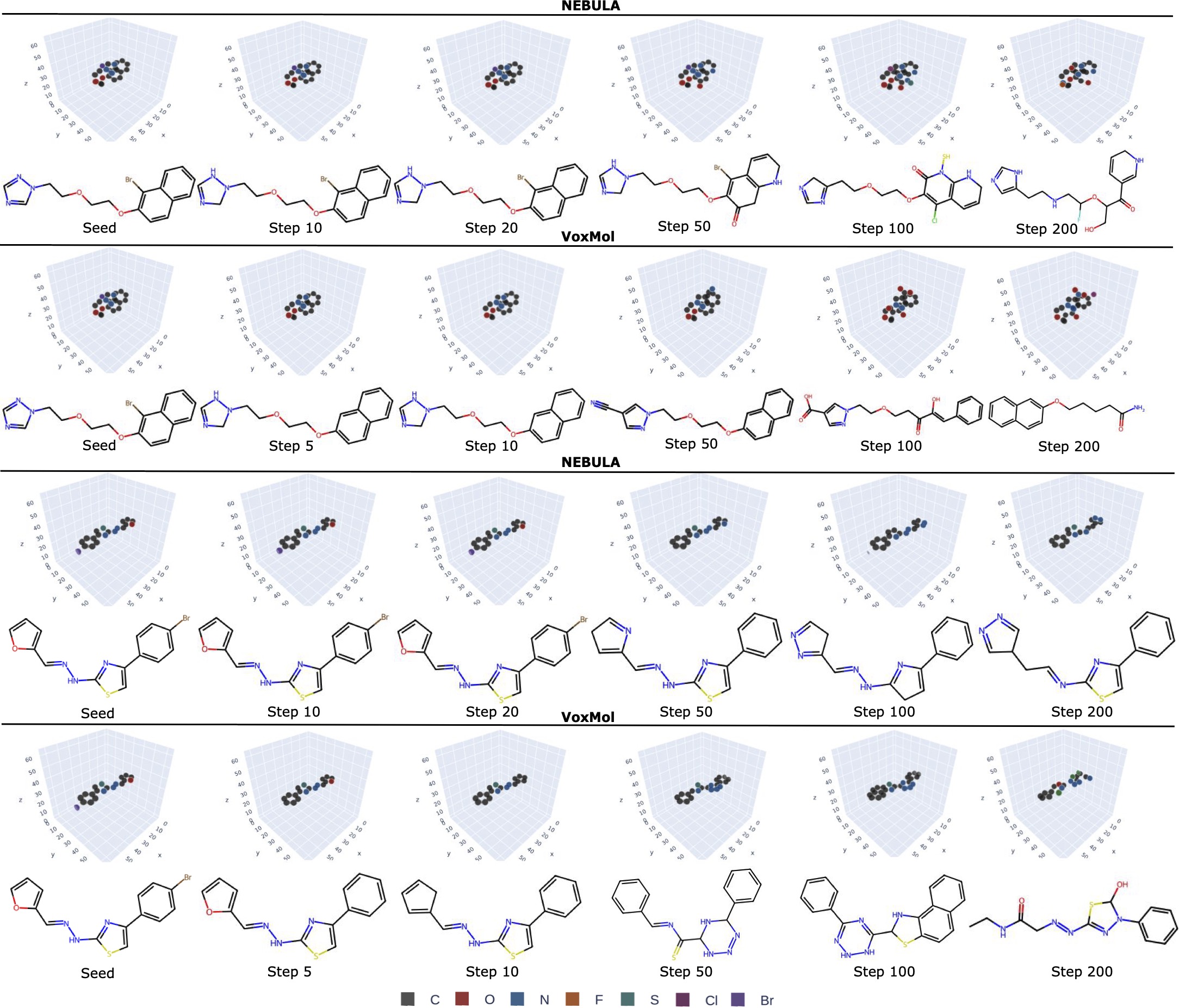}
\caption{\textit{Seeded} generation on \textbf{GEOM} with NEBULA and VoxMol at different WJS steps with the corresponding voxels. Both methods can generate molecules close to the seed in within-dataset generation.}
\label{fig:seeded_drugs}
\vskip -0.2in
\end{figure*}

We show within-dataset results on \textit{seeded} generation where we aim to generate new molecules similar to a provided seed for hypothesis generation. We pass the seed molecule through the VQ-VAE encoder to construct latent embeddings and add noise in the latent space. We generate new molecules by taking different numbers of WJS steps in the noisy latent space and denoising the newly sampled embeddings after the last step. We generate molecules around 1,000 randomly selected seeds from the test set of GEOM (see Table~\ref{tab:seeded_drugs} and Figure~\ref{fig:stab_tan_time} for quantitative results and Figure~\ref{fig:seeded_drugs} for examples of generations). 

Both NEBULA and VoxMol generate stable and valid molecules that are similar to the seed, especially with few WJS steps. We experimentally determined the number of WJS steps for NEBULA and VoxMol that yielded comparable Tanimoto similarity and molecular stability on GEOM. As can be seen, NEBULA maintains significantly higher similarity to seed for much of the duration of WJS trajectories. Though similarity and stability ultimately decay at later steps, it is straightforward to filter unwanted compounds and the efficiency of generation yields overall gains for our task (see Sec~\ref{sec:scalability}).


\textbf{Molecule Sanitization.} Generating molecules via latent WJS requires the removal of skip connections in NEBULA's compression model. We find this makes it more difficult to train the VQ-VAE, in particular to reconstruct the 3D \textit{geometries} of input molecules, as activations are not transferred from encoder to decoder. As a result, while NEBULA generates valid molecules, some heavy atoms are found to be missing attached hydrogens, which leads to an overall low molecular stability metric despite high validity and high atomic stability. Adding missing hydrogens can be solved trivially with fast cheminformatics toolkits such as RDKit~\cite{landrum2016rdkit} via a process known as ``sanitization". In practice, generating molecules with hydrogens is not crucial for discovering new molecules and many models are in fact trained with implicit (removed) hydrogens for efficiency~\cite{vignac2023midi}.

\begin{table*}[h]
\vskip 0.2in
  \begin{tabular}{l | x{25}x{25}x{25}x{25}x{25}x{25}x{25} x{25}x{25} x{25}x{25}}
    WJS Steps & \multirow{1}{*}{\small tan.}  & \multirow{1}{*}{\small stable} & \multirow{1}{*}{\small stable} & \multirow{1}{*}{\small valid}  & \multirow{1}{*}{\small valency} & \multirow{1}{*}{\small atom} & \multirow{1}{*}{\small bond} & \multirow{1}{*}{\small bond} & {\small bond} &  {\small avg. t} \\
    & {\small sim.\;\%$_\uparrow$} & {\small sanit.\;\%$_\uparrow$} & {\small atom\;\%$_\uparrow$} & {\small\%$_\uparrow$} & {\small W$_1$$_\downarrow$} & {\small TV$_\downarrow$} & {\small TV$_\downarrow$} & {\small len\;W$_1$$_\downarrow$} & {\small ang\;W$_1$$_\downarrow$} & {\small [s/mol.]$_\downarrow$} \\
    \shline

    0 (\emph{data}) & 100. & 99.90 &  99.90 & 99.80 & 0.001 & 0.001 & 0.025 & 0.00 & 0.05 & -  \\ 


    \shline
    5 VoxMol & 80.84 \quad\tiny{($\pm$0.96)}  & 85.54 \quad\tiny{($\pm$0.15)} & 99.43 \quad\tiny{($\pm$0.04)} &  90.12 \quad\tiny{($\pm$0.36)} & 0.26 \quad\tiny{($\pm$0)} & \textbf{0.02} \quad\tiny{($\pm$0)} & \textbf{0.03} \quad\tiny{($\pm$0)} & 0.00 \quad\tiny{($\pm$0)}  & 0.67 \quad\tiny{($\pm$0.02)} & 0.38 \\  

    10 & 71.44 \quad\tiny{($\pm$1.82)} & 85.71 \quad\tiny{($\pm$0.36)} & 99.36 \quad\tiny{($\pm$0.04)} &  89.86 \quad\tiny{($\pm$0.28)} & 0.25 \quad\tiny{($\pm$0)} & 0.02 \quad\tiny{($\pm$0)} & 0.03 \quad\tiny{($\pm$0)} & 0.00 \quad\tiny{($\pm$0)}  & 0.67 \quad\tiny{($\pm$0.03)} & 0.66 \\ 

    50 & 44.99 \quad\tiny{($\pm$1.03)}  & 86.42 \quad\tiny{($\pm$0.94)} & 99.35 \quad\tiny{($\pm$0.02)} & 90.15 \quad\tiny{($\pm$0.76)} & 0.25 \quad\tiny{($\pm$0)} & 0.03 \quad\tiny{($\pm$0)} & 0.03 \quad\tiny{($\pm$0)} & 0.00 \quad\tiny{($\pm$0)}  & 0.54 \quad\tiny{($\pm$0.02)} & 0.90 \\ 

    100 & 35.18 \quad\tiny{($\pm$0.34)} & 87.44 \quad\tiny{($\pm$1.06)} & 99.37 \quad\tiny{($\pm$0.04)} & 90.52 \quad\tiny{($\pm$0.81)} & 0.25 \quad\tiny{($\pm$0)} & 0.03 \quad\tiny{($\pm$0)} & 0.03 \quad\tiny{($\pm$0)} & 0.00 \quad\tiny{($\pm$0)}  & \textbf{0.50} \quad\tiny{($\pm$0.00)} & 1.64 \\ 

    200 & 27.37 \quad\tiny{($\pm$0.38)}  & \textbf{88.40} \quad\tiny{($\pm$0.80)} & 99.35 \quad\tiny{($\pm$0.03)} & \textbf{90.86} \quad\tiny{($\pm$0.51)} & 0.25 \quad\tiny{($\pm$0)} & 0.04 \quad\tiny{($\pm$0)} & 0.03 \quad\tiny{($\pm$0)} & 0.00 \quad\tiny{($\pm$0)}  & 0.54 \quad\tiny{($\pm$0.02)} & 3.17 \\ 
    
    \shline

    10 \textbf{NEBULA} & \textbf{88.47} \quad\tiny{($\pm$0.18)}  & 85.73 \quad\tiny{($\pm$0.12)} & \textbf{99.46} \quad\tiny{($\pm$0.01)} &  90.29 \quad\tiny{($\pm$0.11)} & 0.26 \quad\tiny{($\pm$0)} & 0.03 \quad\tiny{($\pm$0)} & \textbf{0.03} \quad\tiny{($\pm$0)} & 0.00 \quad\tiny{($\pm$0)}  & 0.77\quad\tiny{($\pm$0)} & \textbf{0.21} \\  

    20 & 84.3 \quad\tiny{($\pm$0.46)} & 85.46 \quad\tiny{($\pm$0.31)} & 99.42 \quad\tiny{($\pm$0.01)} &  89.82 \quad\tiny{($\pm$0.31)} & 0.26 \quad\tiny{($\pm$0)} & 0.03 \quad\tiny{($\pm$0)} & 0.03 \quad\tiny{($\pm$0)} & 0.00 \quad\tiny{($\pm$0)}  & 0.85 \quad\tiny{($\pm$0.01)} & 0.23 \\  

    50 & 63.85 \quad\tiny{($\pm$0.46)}  & 83.11 \quad\tiny{($\pm$0.31)} & 99.16 \quad\tiny{($\pm$0.01)} &  86.61 \quad\tiny{($\pm$0.31)} & 0.25 \quad\tiny{($\pm$0)} & 0.03 \quad\tiny{($\pm$0)} & 0.03 \quad\tiny{($\pm$0)} & 0.00 \quad\tiny{($\pm$0)}  & 1.15 \quad\tiny{($\pm$0.02)} & 0.28 \\  

    100 & 38.33 \quad\tiny{($\pm$0.8)} & 79.52 \quad\tiny{($\pm$0.85)} & 98.45 \quad\tiny{($\pm$0.04)} &  81.42 \quad\tiny{($\pm$0.75)} & 0.23 \quad\tiny{($\pm$0)} & 0.04 \quad\tiny{($\pm$0)} & 0.03 \quad\tiny{($\pm$0)} & 0.00 \quad\tiny{($\pm$0)}  & 1.80 \quad\tiny{($\pm$0.01)} & 0.36 \\  

    200 & 20.18 \quad\tiny{($\pm$0.29)} & 77.08 \quad\tiny{($\pm$0.64)} & 96.74 \quad\tiny{($\pm$0.02)} & 77.96 \quad\tiny{($\pm$0.52)} & \textbf{0.20} \quad\tiny{($\pm$0)} & 0.07 \quad\tiny{($\pm$0)} & 0.04 \quad\tiny{($\pm$0)} & 0.00 \quad\tiny{($\pm$0)}  & 3.56 \quad\tiny{($\pm$0.06)} & 0.52 \\  

  \end{tabular}
\caption{\textit{Seeded} generation results on \textbf{GEOM}. All results are averaged over 5 repeats of each experiment.}
\label{tab:seeded_drugs}
\vskip -0.2in
\end{table*}

\begin{table*}[h]
\vskip 0.2in
  \begin{tabular}{l | x{25}x{25}x{25}x{25}x{25}x{25}x{25} x{25}x{40} x{40}}
   WJS Steps & \multirow{1}{*}{\small tan.} & \multirow{1}{*}{\small stable} &  \multirow{1}{*}{\small stable} & \multirow{1}{*}{\small valid}  & \multirow{1}{*}{\small valency} & \multirow{1}{*}{\small atom} & \multirow{1}{*}{\small bond} & \multirow{1}{*}{\small bond len} & \multirow{1}{*}{\small bond ang}  \\
    & {\small sim.\;\%$_\uparrow$} & {\small sanit.\;\%$_\uparrow$} & {\small atom\;\%$_\uparrow$} & {\small\%$_\uparrow$} & {\small W$_1$$_{\downarrow\Delta}$} & {\small TV$_{\downarrow\Delta}$} & {\small TV$_{\downarrow\Delta}$} & {\small \;W$_1$$_{\downarrow\Delta}$} & {\small \;W$_1$$_{\downarrow\Delta}$}  \\
    \shline

    0 (\emph{data}) & 100.00 & 95.51 & 99.62 & 99.40 & 2.98 & 0.90 & 0.24 & 0.01 & 3.40 \\ 

    \shline
    
    5 VoxMol & 10.95 \quad\tiny{($\pm$1.85)} & 92.45 \quad\tiny{($\pm$0.29)} & 95.60 \quad\tiny{($\pm$0.07)} & 97.11 \quad\tiny{($\pm$0.17)} & 0.21 \quad\tiny{($\pm$0)} & 0.43 \quad\tiny{($\pm$0.01)} & 0.14 \quad\tiny{($\pm$0)} & 0.00 \quad\tiny{($\pm$0)}  & 4.04 \quad\tiny{($\pm$0.07)} \\  

    10 & 9.74 \quad\tiny{($\pm$0.62)} & 92.08 \quad\tiny{($\pm$0.83)} & 96.89 \quad\tiny{($\pm$0.11)} & 96.49 \quad\tiny{($\pm$0.63)} & 0.22 \quad\tiny{($\pm$0)} & 0.35 \quad\tiny{($\pm$0.00)} & 0.14 \quad\tiny{($\pm$0)} & 0.00 \quad\tiny{($\pm$0)}  & \textbf{3.49} \quad\tiny{($\pm$0.05)}  \\  

    50 & 9.75 \quad\tiny{($\pm$0.38)} & 90.73 \quad\tiny{($\pm$0.54)} & 98.44 \quad\tiny{($\pm$0.07)} & 95.17 \quad\tiny{($\pm$0.30)} & 0.24 \quad\tiny{($\pm$0)} & 0.15 \quad\tiny{($\pm$0.01)} & 0.13 \quad\tiny{($\pm$0)} & 0.00 \quad\tiny{($\pm$0)}  & 2.36 \quad\tiny{($\pm$0.04)}  \\  

    100 & 9.81 \quad\tiny{($\pm$0.21)} & 89.93 \quad\tiny{($\pm$0.47)} & 98.62 \quad\tiny{($\pm$0.04)} & 94.81 \quad\tiny{($\pm$0.43)} & 0.25 \quad\tiny{($\pm$0)} & 0.10 \quad\tiny{($\pm$0.01)} & 0.12 \quad\tiny{($\pm$0)} & 0.00 \quad\tiny{($\pm$0)}  & 1.98 \quad\tiny{($\pm$0.07)}  \\  

    200 & 9.86 \quad\tiny{($\pm$0.22)} & 90.44 \quad\tiny{($\pm$0.85)} & \textbf{98.77} \quad\tiny{($\pm$0.07)} & 94.68 \quad\tiny{($\pm$0.56)} & \textbf{0.25} \quad\tiny{($\pm$0)} & 0.06 \quad\tiny{($\pm$0)} & 0.12 \quad\tiny{($\pm$0)} & 0.00 \quad\tiny{($\pm$0)}  & 1.62 \quad\tiny{($\pm$0.05)}  \\  

    \shline

    10 \textbf{NEBULA} & \textbf{35.1} \quad\tiny{($\pm$0.57)}  & 95.18 \quad\tiny{($\pm$0.26)} & 75.64 \quad\tiny{($\pm$0.13)} &  \textbf{98.86} \quad\tiny{($\pm$0.23)} & 0.23 \quad\tiny{($\pm$0)} & \textbf{0.84} \quad\tiny{($\pm$0)} & \textbf{0.24} \quad\tiny{($\pm$0)} & \textbf{0.01} \quad\tiny{($\pm$0)}  & 4.59 \quad\tiny{($\pm$0.03)} \\  

    20 & 32.19 \quad\tiny{($\pm$0.38)}  & 95.44 \quad\tiny{($\pm$0.27)} & 76.43 \quad\tiny{($\pm$0.24)} &  98.80 \quad\tiny{($\pm$0.13)} & 0.22 \quad\tiny{($\pm$0)} & 0.83 \quad\tiny{($\pm$0)} & 0.23 \quad\tiny{($\pm$0)} & 0.01 \quad\tiny{($\pm$0)}  & 4.72 \quad\tiny{($\pm$0.05)}  \\  

    50 & 25.62 \quad\tiny{($\pm$0.26)}  & 95.64 \quad\tiny{($\pm$0.26)} & 78.35 \quad\tiny{($\pm$0.18)} &  98.36 \quad\tiny{($\pm$0.12)} & 0.20 \quad\tiny{($\pm$0)} & 0.79 \quad\tiny{($\pm$0)} & 0.23 \quad\tiny{($\pm$0)} & 0.01 \quad\tiny{($\pm$0)}  & 5.19 \quad\tiny{($\pm$0.06)}  \\  

    100 & 18.38 \quad\tiny{($\pm$0.27)}  & \textbf{96.16} \quad\tiny{($\pm$0.19)} & 80.04 \quad\tiny{($\pm$0.20)} & 97.88 \quad\tiny{($\pm$0.35)} & 0.18 \quad\tiny{($\pm$0)} & 0.74 \quad\tiny{($\pm$0)} & 0.23 \quad\tiny{($\pm$0)} & 0.01 \quad\tiny{($\pm$0)}  & 6.39 \quad\tiny{($\pm$0.04)}  \\  

    200 & 12.58 \quad\tiny{($\pm$0.09)}  & 95.72 \quad\tiny{($\pm$0.49)} & 79.77 \quad\tiny{($\pm$0.51)} & 96.72 \quad\tiny{($\pm$0.26)} & 0.19 \quad\tiny{($\pm$0.01)} & 0.68 \quad\tiny{($\pm$0)} & 0.23 \quad\tiny{($\pm$0)} & 0.00 \quad\tiny{($\pm$0)}  & 9.30 \quad\tiny{($\pm$0.09)}  \\  
  \end{tabular}
\caption{Cross-dataset generalizability: \textit{seeded} generation results on \textbf{PCQM}. NEBULA is able to generate molecules with atom and bond distributions closer to the new dataset (step 0). Bolded distribution metrics are those closest to the seed ($\downarrow\Delta$). All results are averaged over 5 repeats of each experiment.}
\label{tab:seeded_pubchem}
\vskip -0.2in
\end{table*}

\subsection{Scalability}\label{sec:scalability}

\TODO{update this subsection with final results}



Sampling in NEBULA's latent space is on average 6 times faster than the baselines (see the last column of Table~\ref{tab:seeded_drugs} and in Figure~\ref{fig:stab_tan_time}) allowing us to quickly generate large molecular libraries with hundreds of thousands of molecules. Drug-like molecules require a voxel grid size of at least $[8 \times 64 \times 64 \times 64]$, where 8 is the number of atom types, requiring $\sim$ 2 million points to represent each molecule and to sample new molecules in the input voxel space as is done in VoxMol. In contrast, NEBULA compresses the voxel space to a much smaller latent representation of $[1024 \times 8 \times 8 \times 8]$, where 1024 is the latent dimension, requiring only $\sim$ 500 k points per molecule which is 4 times smaller than the input voxel space representation. As can be seen in Figure~\ref{fig:stab_tan_time}, NEBULA trajectories require very minimal compute time, even at higher WJS steps.



\TODO{show results with 64-latent dimension - faster but less stable and may not generalize as well to new datasets}

\subsection{Cross-Dataset Generalizability}
\begin{figure*}
\vskip 0.2in
\centering
    \includegraphics[width=0.98\textwidth]{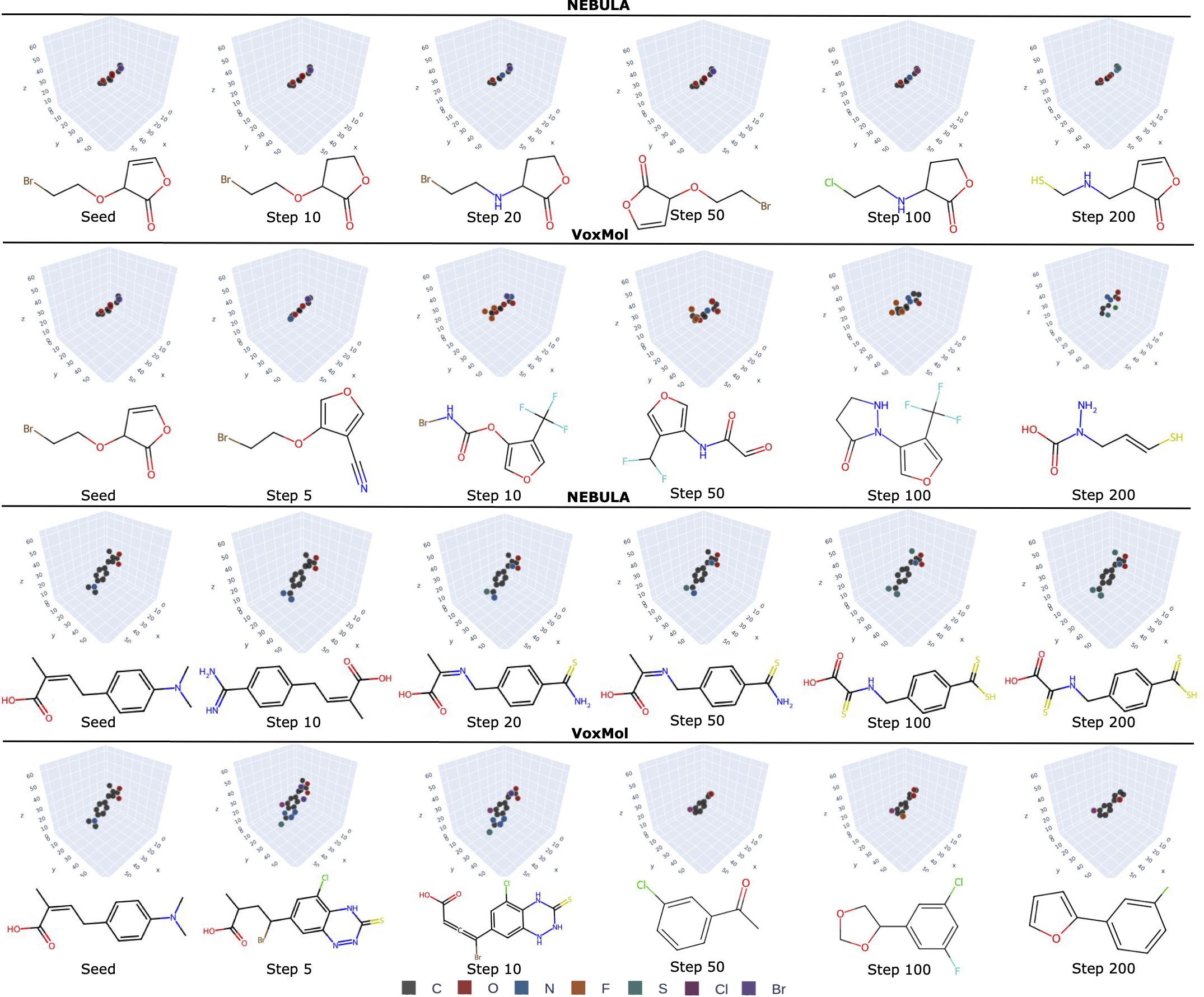}
\caption{\textit{Seeded} generation on \textbf{PCQM} with NEBULA and VoxMol at different WJS steps with the corresponding voxels. NEBULA is able to generate molecules that maintain the seed scaffold in all cases in cross-dataset generation, while VoxMol tends to diverge from the seed compound.}
\label{fig:seeded_PubChem}

\vskip -0.2in
\end{figure*}

We compare the cross-dataset generalizability of NEBULA and VoxMol trained on GEOM by generating libraries around 1,000 randomly selected seeds from the PCQM test set (see Table~\ref{tab:seeded_pubchem} and Figure~\ref{fig:stab_tan_time} for quantitative results and Figure~\ref{fig:seeded_PubChem} for example generations). NEBULA generates molecules that are much closer to the seed at all WJS steps as demonstrated by \textit{\textbf{tan. sim. \%}}. VoxMol tends to generate molecules which are quite different from PCQM, even at few WJS steps. Figure~\ref{fig:seeded_PubChem} shows that NEBULA makes incremental changes to seeds at early steps and maintains their overall scaffolds through entire trajectories, while VoxMol tends to generate very different molecule not resembling the seed. NEBULA also generates molecules with \textit{\textbf{atom TV}} and \textit{\textbf{bond TV}} that are very close to the raw PCQM metrics (step 0), while VoxMol seems to be reverting back to molecules that are more similar to its training distribution, GEOM.

We hypothesize that NEBULA achieves robustness to new chemical spaces due to the removal of skip connections in the VQ-VAE (see Sec~\ref{sec:data_and_methods}). While this makes MSE loss slower to converge, it may also prevent ``shortcuts" such as fitting geometric features that are specific to a certain structure-computation method but that do not generalize~\cite{maser2023moleclues}. A similar behavior was observed for U-Net architectures used in image segmentation~\cite{wilm2024rethinking, kamath2023we}. This is visible in \textit{\textbf{bond ang W}}$_1$, which shows the distribution of bond angles  diverges from the training set at long WJS steps for NEBULA but not for VoxMol. Future work will seek to identify training and data strategies that maximize chemical-space robustness~\cite{maser2023supsiam} while also producing faithful geometries.

\begin{figure*}
\vskip 0.2in
    \includegraphics[width=0.98\textwidth]{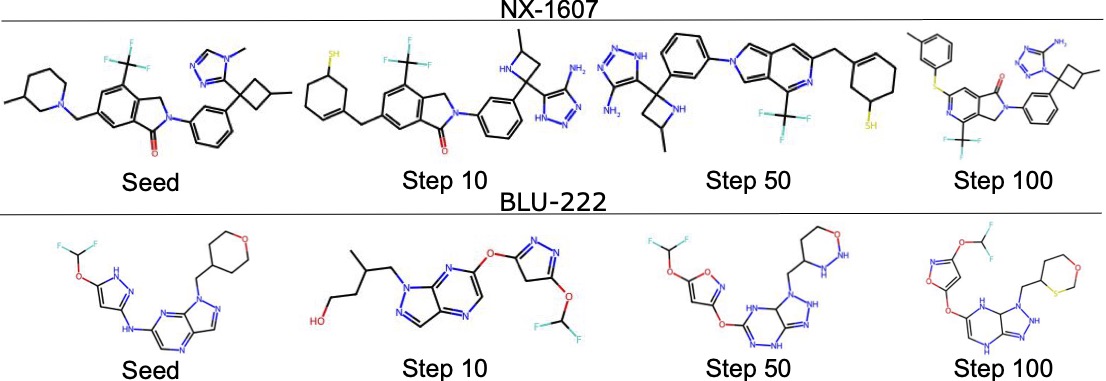}
\caption{\textit{Seeded} Generation on real drugs just released in March 2024~\cite{ACS} with NEBULA at different WJS steps.}
\label{fig:seeded_ACS}
\vskip -0.2in
\end{figure*}


\textbf{Real-world Application.} We additionally demonstrate that NEBULA can generalize to unseen real drug molecules that were recently publicly released (March 2024)~\cite{ACS}. These molecules represent significant leaps in complexity from GEOM and PCQM and have real biological relevance; all are currently in clinical trials as advanced cancer therapeutics. While similarity and stability measures are lower than for other datasets (Table~\ref{tab:seeded_ACS}), NEBULA is able to generate molecules that remain very faithful to the seed scaffold and introduce functional-group changes that typify a medicinal chemistry process (Figure~\ref{fig:seeded_ACS}).

\begin{table}
\vskip 0.2in
  \begin{tabular}{y{22}|x{20}x{22}x{20}x{20}x{20}}
    {\small WJS}  & \multirow{1}{*}{\small tan.} & \multirow{1}{*}{\small stable} & \multirow{1}{*}{\small stable} & \multirow{1}{*}{\small valid}   \\
    {\small Steps} & {\small sim.\;\%$_\uparrow$} & {\small sanit.\;\%$_\uparrow$} & {\small atom\;\%$_\uparrow$} & {\small\%$_\uparrow$}  \\
    \shline
   
    

10 & 30.62 & 42.22 & 42.20 & 42.22 \\
20 & 31.13 & 40.00 & 42.44 & 40.00 \\ 
50 &23.37 & 43.33 & 42.93 & 43.33 \\
100 & 22.95 & 42.22 & 43.48 & 42.22 \\
200 & 19.17 & 43.33 & 43.26 & 43.33 \\

    \vskip -0.5in
\caption{\textit{Seeded} generation results averaged over 10 experiments for 5 real recently released drugs~\cite{ACS} (for a total of 50 generations).}
\label{tab:seeded_ACS}
\end{tabular}
\vskip -0.2in
\end{table}

See the Appendix for more examples of qualitative generation results on GEOM, PCQM, and drug molecules~\cite{ACS}.

\section{Conclusions}
We present the first latent generative model based on 3D-voxel representations of molecules, called NEBULA. NEBULA produces high quality, stable, and valid molecules around a seed molecule and is scalable to generation of very large molecular libraries needed for drug discovery via efficient \textit{walk-jump} sampling in latent space. Moreover, NEBULA generalizes well to new chemical spaces, including fragments and real drug molecules. We expect to see great utility of our approach for accelerating ML-based drug discovery.

\clearpage

\bibliography{example_paper}
\bibliographystyle{icml2024}

\newpage
\appendix
\onecolumn

\section{Additional Implementation Details}

\subsection{Training and Architecture Details}

We use a VQ-VAE 3D autoencoder with no skip connections as the compression model architecture and a 3D U-Net with skip connections for the latent denoiser. Both models use a 3D convolutional architecture similar to~\cite{pinheiro20233d}, with 4 levels of resolution and self-attention on the lowest two resolutions. We found that a latent dimension of 1024 worked best with higher noise levels. We reduce the 1024 latent embedding to 32 in the first layer of the latent denoiser and gradually increase it with subsequent layers. We use a commitment cost of 0.25 and 256 latent codebook embeddings of 256 for the compression VQ-VAE. We train the compression and latent denoising models for about 150 epochs on GEOM-Drugs (until the mean intersection over union between the input and reconstructed voxels is above 0.90). We train the models with a batch size 32, AdamW optimizer~\cite{loshchilov2017decoupled} with $\beta_{1}$=0.9, $\beta_{2}$=0.999, learning rate $10^{-5}$, weight decay $10^{-2}$, dropout 0.1, SiLU activation function, and we update the weights with exponential moving average (EMA) with decay of 0.999. We augment all models during training by randomly rotating and translating every training sample. We use a noise level of $\sigma=1.8$ for all generations, friction of 1.0, lipschitz of 1.0 and step size of 0.25 (VoxMol uses step size of 0.5 in the voxel space) for the MCMC sampling to generate new molecules with the trained latent denoising model.



\subsection{Evaluation Metrics} We evaluate the models using the metrics following~\cite{vignac2023midi}. \textit{\textbf{stable mol}} and \textit{\textbf{stable atom}} are molecular and atomic stability; an atom is stable when the number of bonds with other atoms matches their valence and a molecule is stable only if 100 \% of its atoms are stable~\cite{hoogeboom2022equivariant} (see Sec~\ref{sec:molecular_library_generation}). \textit{\textbf{validity}} is measured as the percent of molecules passing the RDKit's~\cite{landrum2016rdkit} sanitization. \textit{\textbf{valency W$_1$}} is the Wasserstein distances between valences in the distributions of the generated molecules and the molecules in the dataset. \textit{\textbf{atoms TV}} and \textit{\textbf{bonds TV}} are the total variation between the atom and bond types distributions. \textit{\textbf{bond length W$_1$}} and \textit{\textbf{bond angle W$_1$}} are the Wasserstein distances between the distributions of bond lengths and angles of the generated molecules and the molecules in the dataset.

\clearpage

\section{Additional Generation Results (Within-Dataset)}

\begin{figure*}[h]
\vskip 0.2in
\begin{center}
\centerline{\includegraphics[width=\columnwidth]{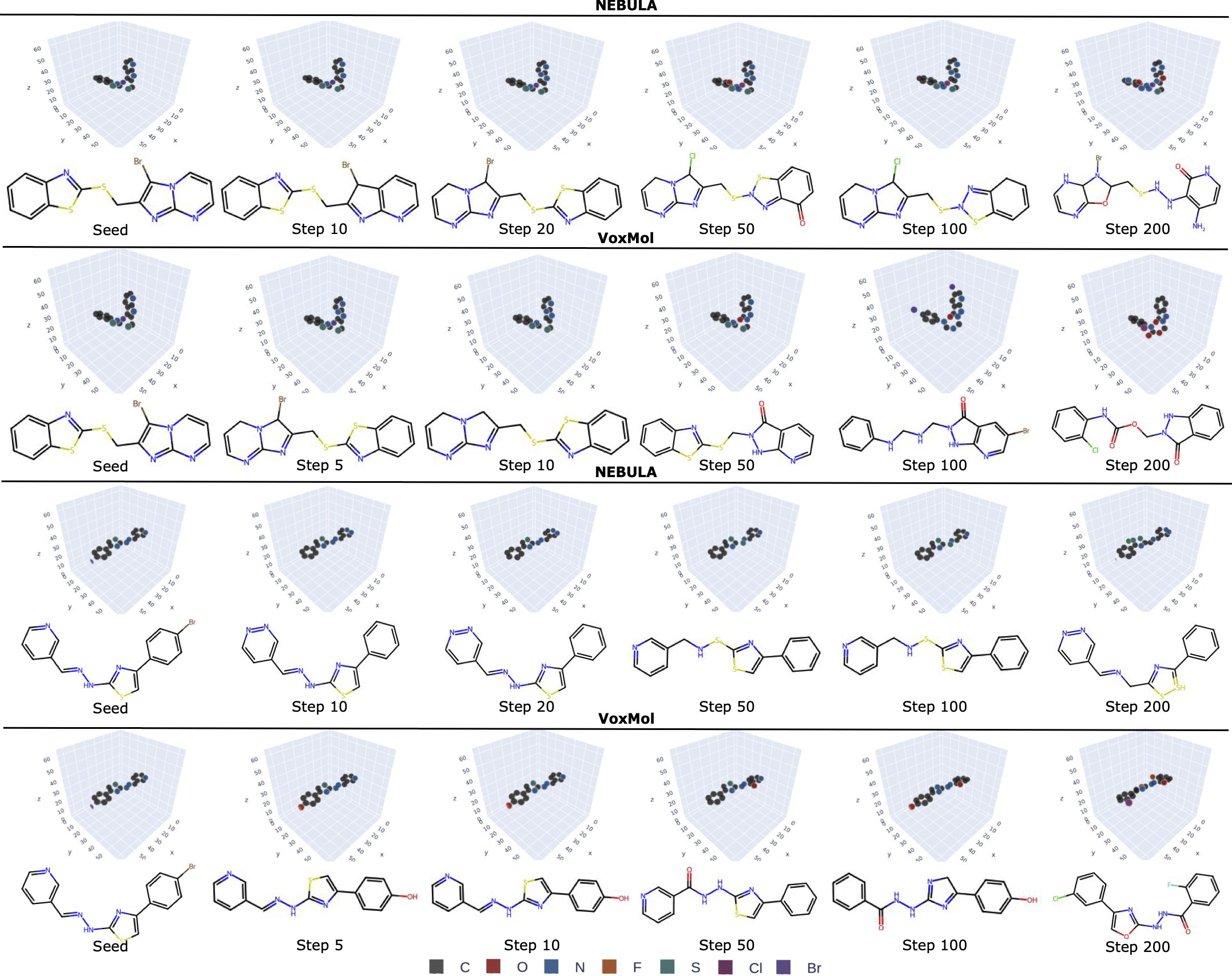}}
\caption{Additional examples of seeded generation on GEOM-Drugs~\cite{geomdrugs_axelrod2022geom}.}
\label{fig:appendix_drugs_nebula1}
\end{center}
\vskip -0.2in
\end{figure*}

\begin{figure*}[h]
\vskip 0.2in
\begin{center}
\centerline{\includegraphics[width=\columnwidth]{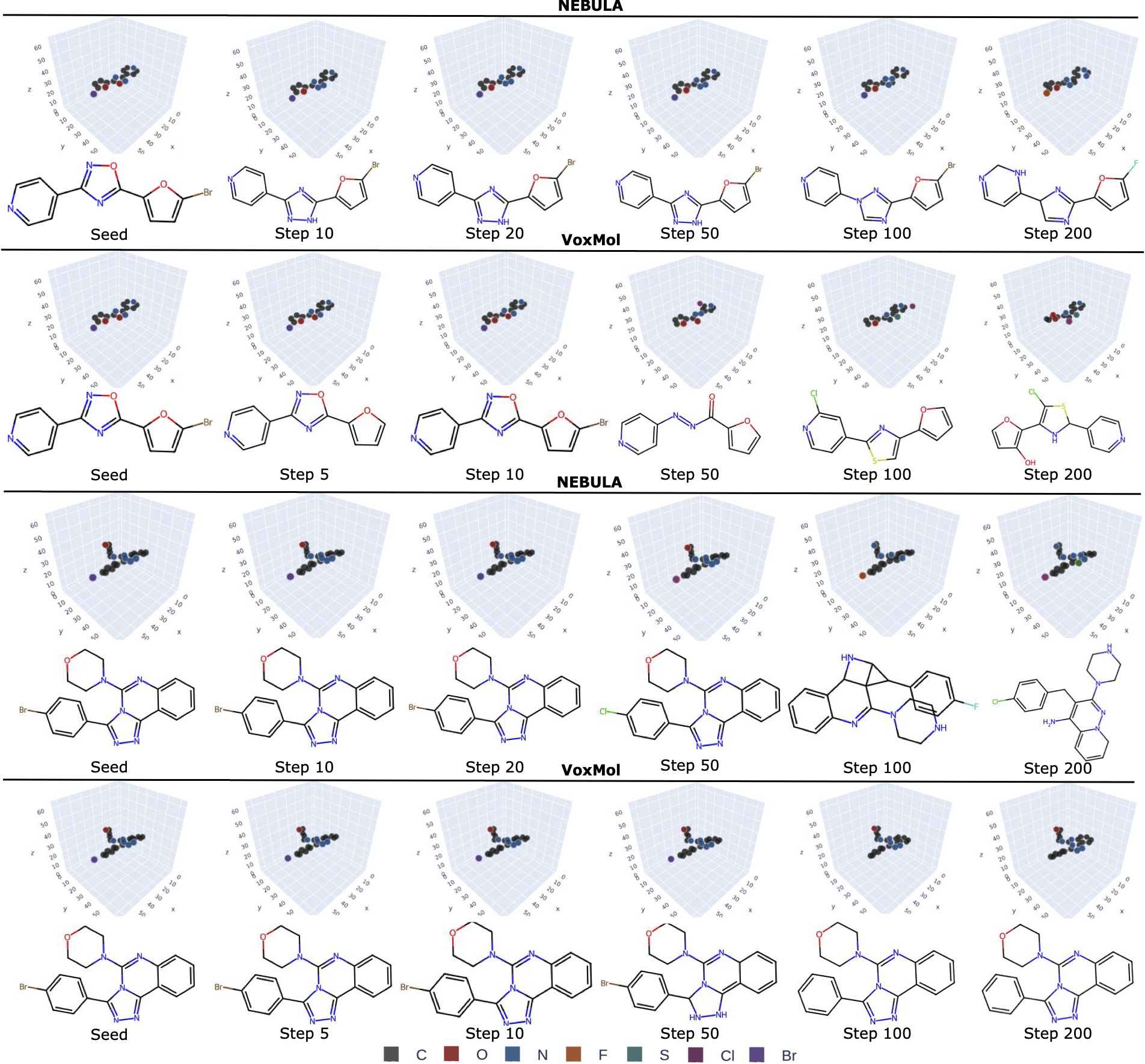}}
\caption{Additional examples of seeded generation on GEOM-Drugs~\cite{geomdrugs_axelrod2022geom}.}
\label{fig:appendix_drugs_nebula2}
\end{center}
\vskip -0.2in
\end{figure*}

\begin{figure*}[h]
\vskip 0.2in
\begin{center}
\centerline{\includegraphics[width=\columnwidth]{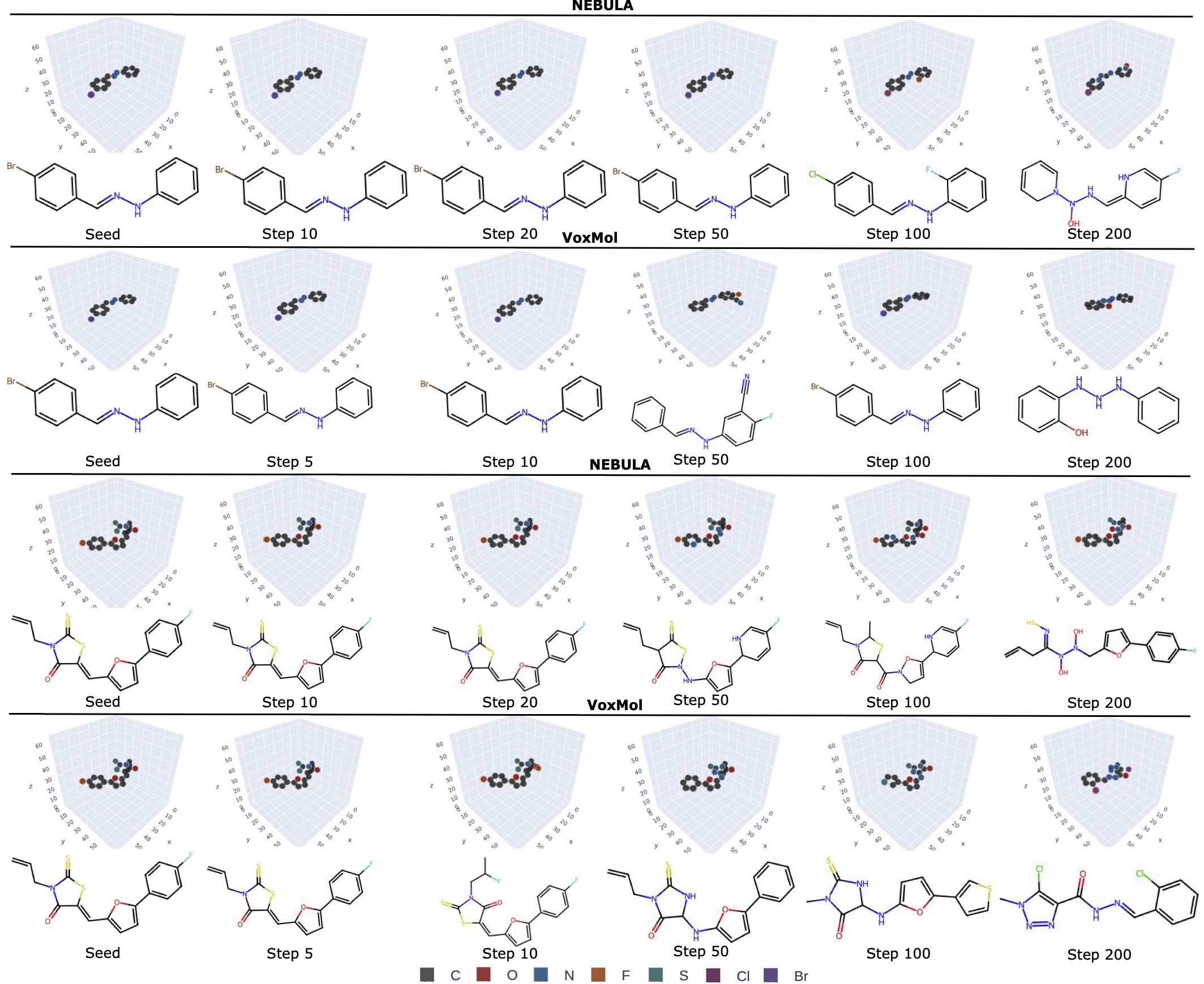}}
\caption{Additional examples of seeded generation on GEOM-Drugs~\cite{geomdrugs_axelrod2022geom}.}
\label{fig:appendix_drugs_nebula3}
\end{center}
\vskip -0.2in
\end{figure*}

\begin{figure*}[h]
\vskip 0.2in
\begin{center}
\centerline{\includegraphics[width=\columnwidth]{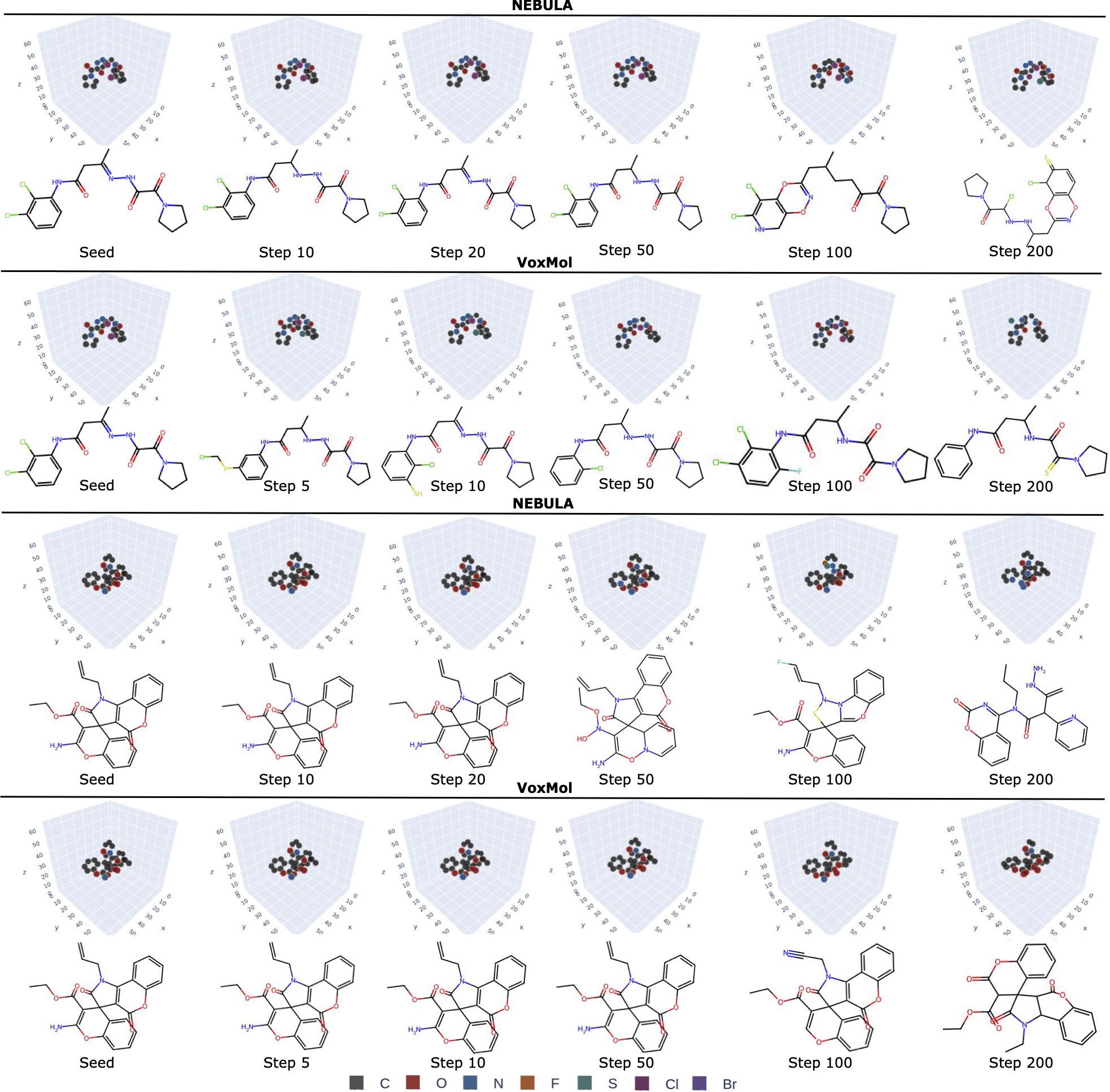}}
\caption{Additional examples of seeded generation on GEOM-Drugs~\cite{geomdrugs_axelrod2022geom}.}
\label{fig:appendix_drugs_nebula4}
\end{center}
\vskip -0.2in
\end{figure*}

\clearpage

\section{Additional Generation Results (Cross-Dataset)}

\begin{figure*}[h]
\vskip 0.2in
\begin{center}
\centerline{\includegraphics[width=\columnwidth]{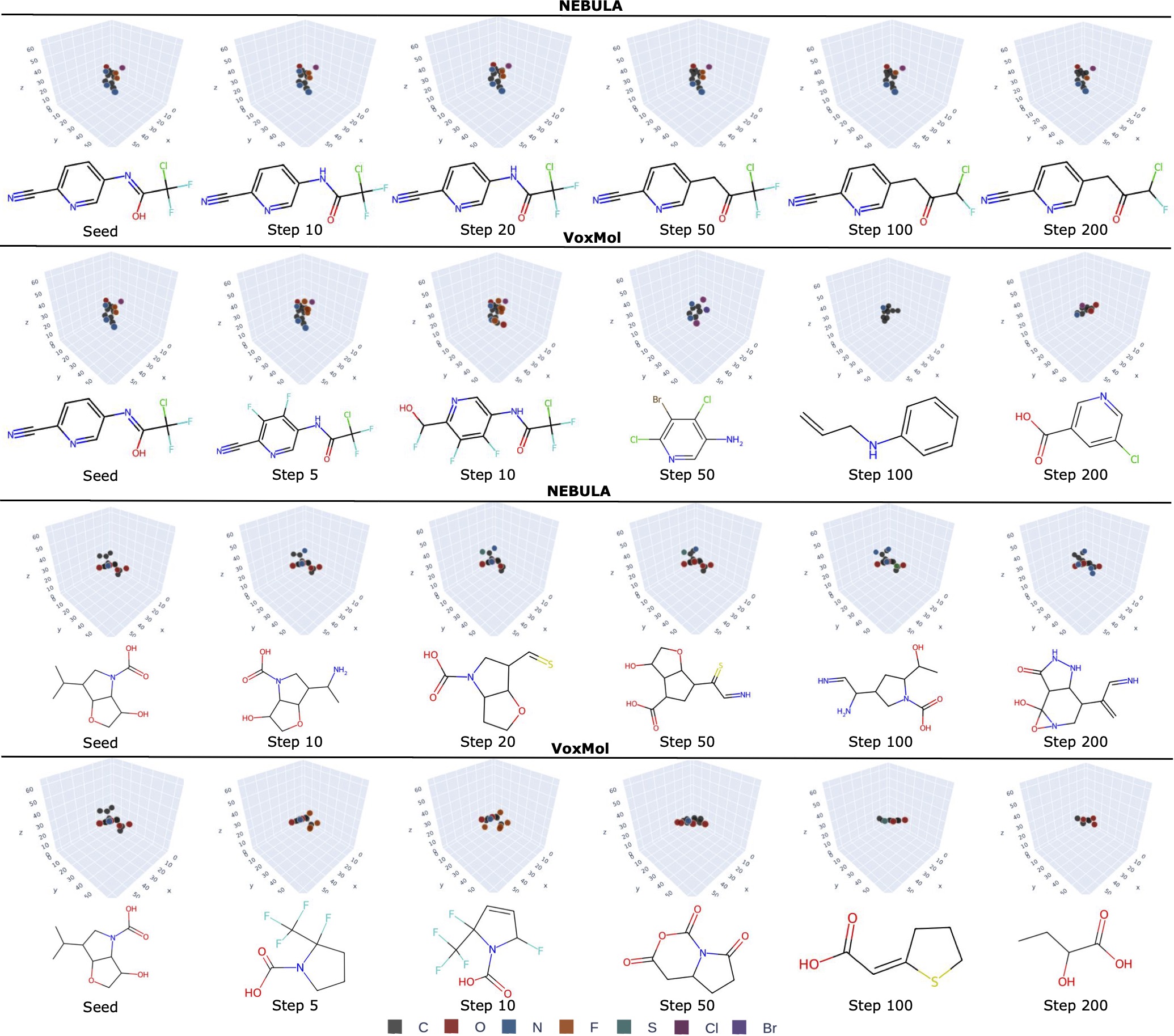}}
\caption{Additional examples of seeded generation on PubChem~\cite{pubchem1}.}
\label{fig:appendix_pubchem_nebula1}
\end{center}
\vskip -0.2in
\end{figure*}

\begin{figure*}[h]
\vskip 0.2in
\begin{center}
\centerline{\includegraphics[width=\columnwidth]{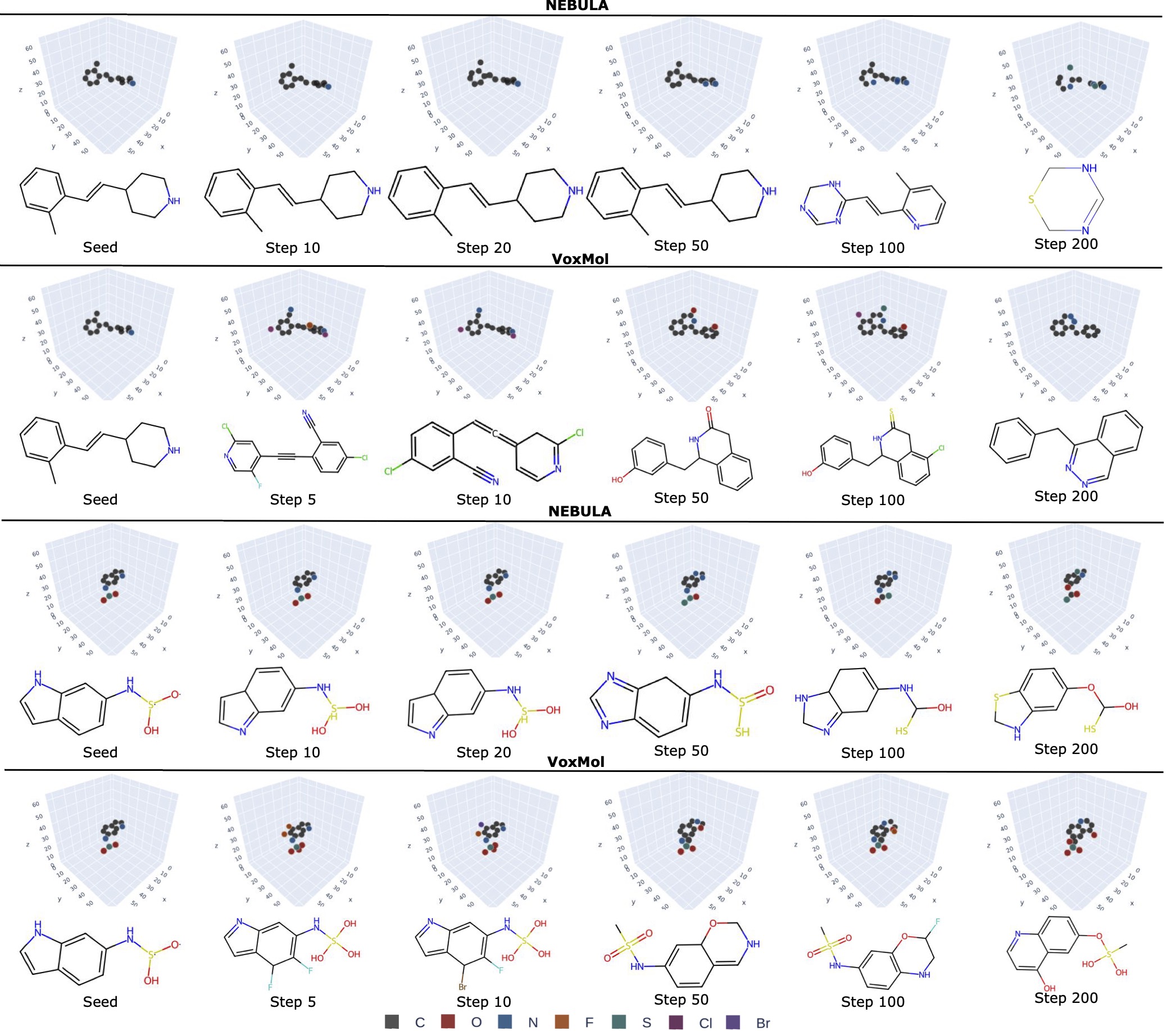}}
\caption{Additional examples of seeded generation on PubChem~\cite{pubchem1}.}
\label{fig:appendix_pubchem_nebula2}
\end{center}
\vskip -0.2in
\end{figure*}

\begin{figure*}[h]
\vskip 0.2in
\begin{center}
\centerline{\includegraphics[width=\columnwidth]{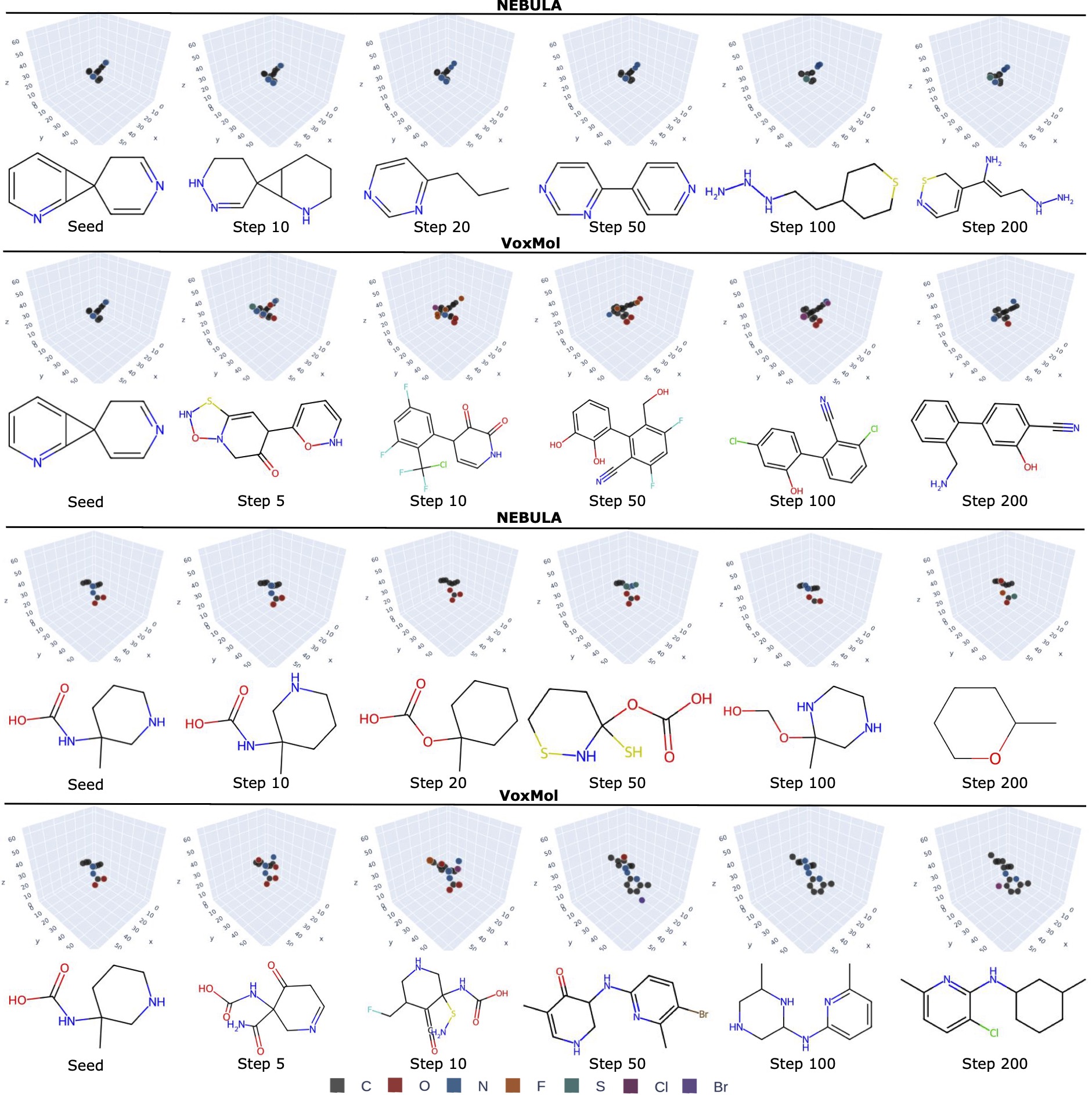}}
\caption{Additional examples of seeded generation on PubChem~\cite{pubchem1}.}
\label{fig:appendix_pubchem_nebula3}
\end{center}
\vskip -0.2in
\end{figure*}

\begin{figure*}[h]
\vskip 0.2in
\begin{center}
\centerline{\includegraphics[width=\columnwidth]{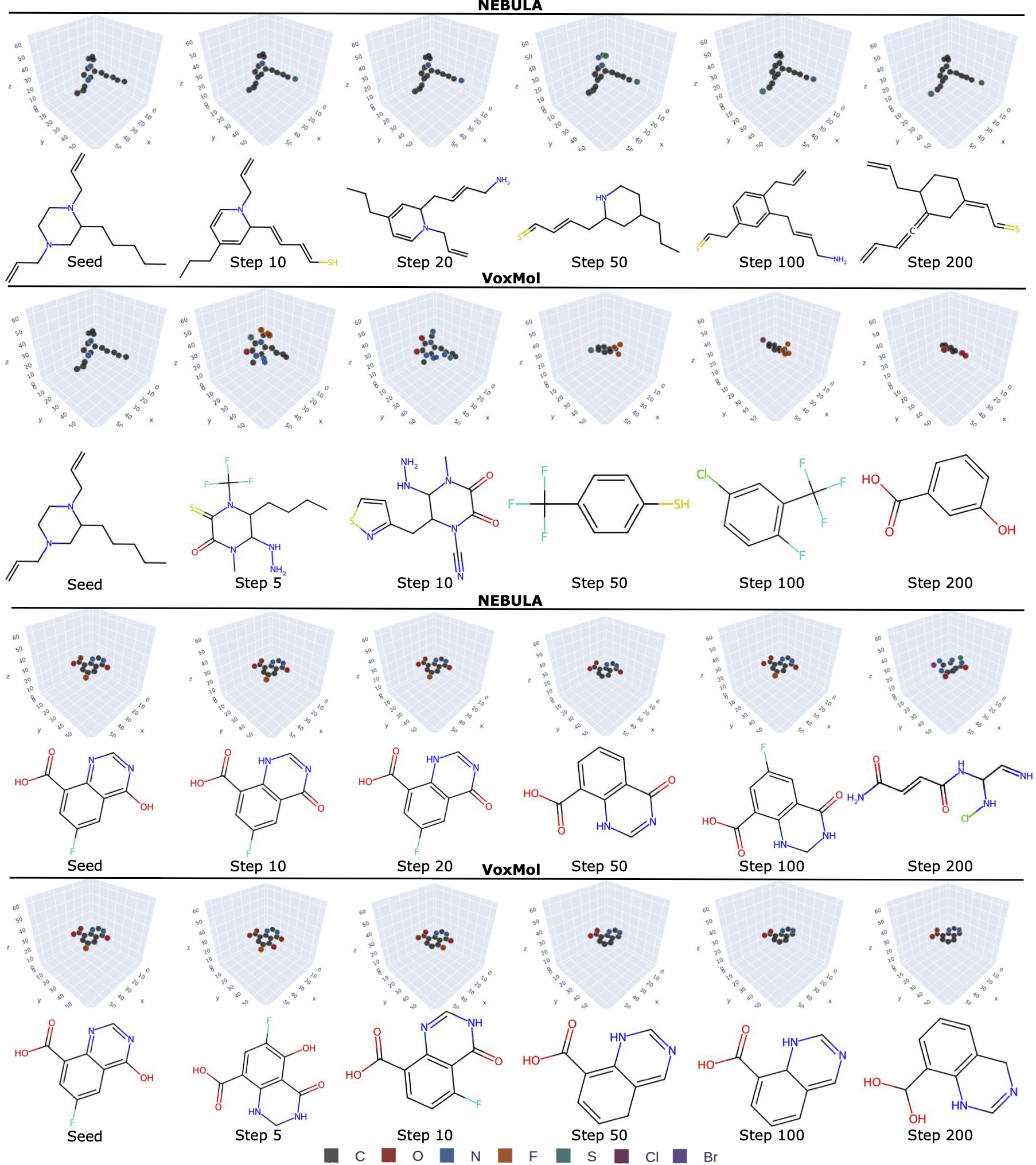}}
\caption{Additional examples of seeded generation on PubChem~\cite{pubchem1}.}
\label{fig:appendix_pubchem_nebula4}
\end{center}
\vskip -0.2in
\end{figure*}

\begin{figure*}[h]
\vskip 0.2in
\begin{center}
\centerline{\includegraphics[width=\columnwidth]{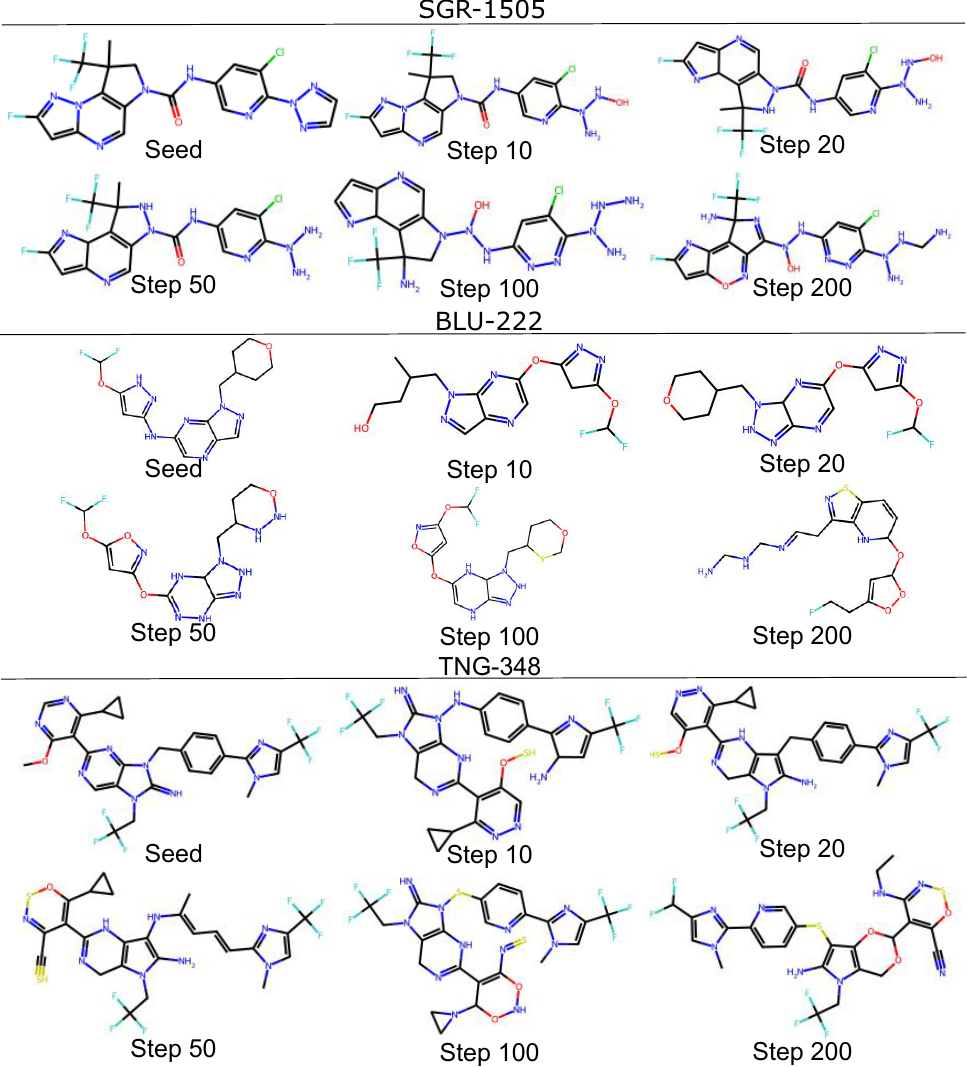}}
\caption{Additional examples of seeded generation on recently released drugs~\cite{ACS} with additional WJS steps.}
\label{fig:appendix_ACS1}
\end{center}
\vskip -0.2in
\end{figure*}

\begin{figure*}[h]
\vskip 0.2in
\begin{center}
\centerline{\includegraphics[width=\columnwidth]{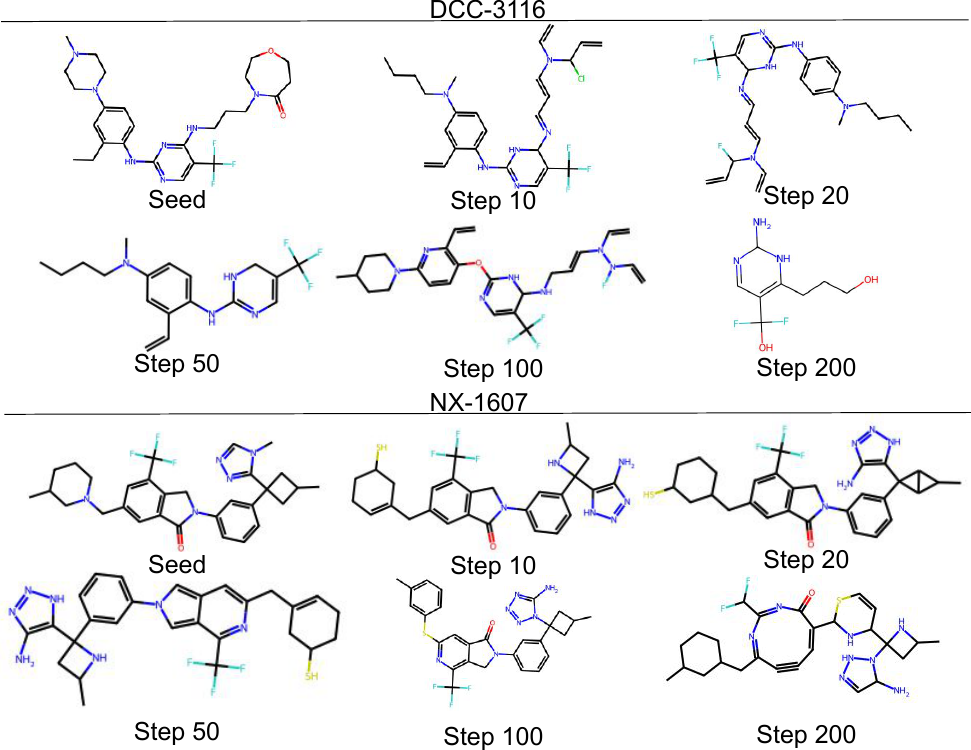}}
\caption{Additional examples of seeded generation on recently released drugs~\cite{ACS} with additional WJS steps.}
\label{fig:appendix_ACS2}
\end{center}
\vskip -0.2in
\end{figure*}

\end{document}